\documentclass[11pt,a4paper]{article}
\usepackage[hyperref]{acl2021}
\usepackage{times}
\usepackage{latexsym}
\usepackage{amsmath, amssymb}
\usepackage{graphicx}

\usepackage{microtype}

\aclfinalcopy 
\def\aclpaperid{603} 

\usepackage{pgf}
\usepackage{tikz}
\usepackage{pgfplots}
\usepackage{ifthen}
\usepackage{subcaption}
\usepackage{booktabs}
\usetikzlibrary{matrix,calc}
\usepackage{amsmath}

\newcommand{\ignore}[1]{}

\title{Joint Verification and Reranking for Open Fact Checking Over Tables}

\author{Michael Schlichtkrull\textsuperscript{1}\thanks{\hspace{.06in}Work done while interning with Facebook AI Research.}, Vladimir Karpukhin\textsuperscript{2}, Barlas Oğuz\textsuperscript{2},\\ \textbf{Mike Lewis\textsuperscript{2}, Wen-tau Yih\textsuperscript{2}, Sebastian Riedel\textsuperscript{2,3}} \\
\textsuperscript{1}University of Cambridge, 
\textsuperscript{2}Facebook AI Research,
\textsuperscript{3}University College London \\
{\tt michael.schlichtkrull@cst.cam.ac.uk,}\\
{\tt \{vladk, barlaso, mikelewis, scottyih, sriedel\}@fb.com}}

\date{}

\begin{document}
\maketitle
\begin{abstract}
Structured information is an important knowledge source for automatic verification of factual claims. Nevertheless, the majority of existing research into this task has focused on textual data, and the few recent inquiries into structured data have been for the closed-domain setting where appropriate evidence for each claim is assumed to have already been retrieved. In this paper, we investigate verification over structured data in the open-domain setting, introducing a joint reranking-and-verification model which fuses evidence documents in the verification component. Our open-domain model achieves performance comparable to the closed-domain state-of-the-art on the TabFact dataset, and demonstrates performance gains from the inclusion of multiple tables as well as a significant improvement over a heuristic retrieval baseline.
\end{abstract}
\section{Introduction}

Verifying whether a given fact coheres with a trusted body of knowledge is a fundamental problem in NLP, with important applications to automated fact checking~\citep{vlachos-riedel-2014-fact} and other tasks in computational journalism~\citep{cohen2011, flew2012promise}. Despite extensive investigation of the problem under different conditions including entailment and natural language inference~\citep{dagan2005pascal, bowman-etal-2015-large} as well as claim verification ~\citep{vlachos-riedel-2014-fact, alhindi2018your, thorne-vlachos-2018-automated}, relatively little attention has been devoted to the setting where the trusted body of evidence is structured in nature --- that is, where it consists of tabular or graph-structured data. 

Recently, two datasets were introduced for claim verification over tables~\citep{chen2019tabfact, gupta-etal-2020-infotabs}. 
In both datasets, claims can be verified given a single associated table. 
While highly useful for the development of models, this \textit{closed} setting is not reflective of real-world fact checking tasks where it is usually not known which table to consult for evidence. Realistic systems must first retrieve evidence from a large data source. That is, realistic systems must operate in an \textit{open} setting.

\begin{figure}[t]
    \centering
    \includegraphics[width=0.35\textwidth]{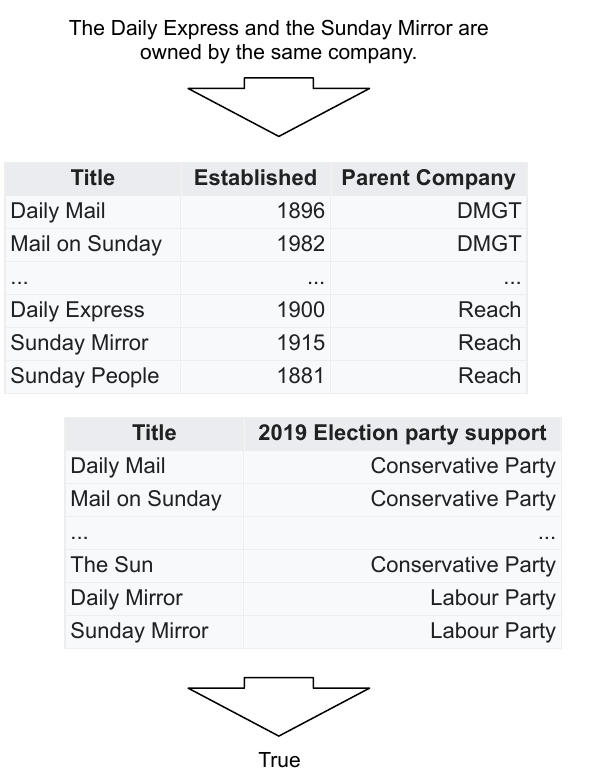}
    \hspace{-0.2cm}\vspace{-0.1cm} 
    \caption{Example query to be evaluated against two retrieved tables. Named entities represent a strong baseline for retrieval, but ultimately a more complex model is required to distinguish highly similar tables.}
    \label{fig:example_query}
\end{figure}

\begin{figure*}[t]
    \centering
    \includegraphics[width=0.97\textwidth]{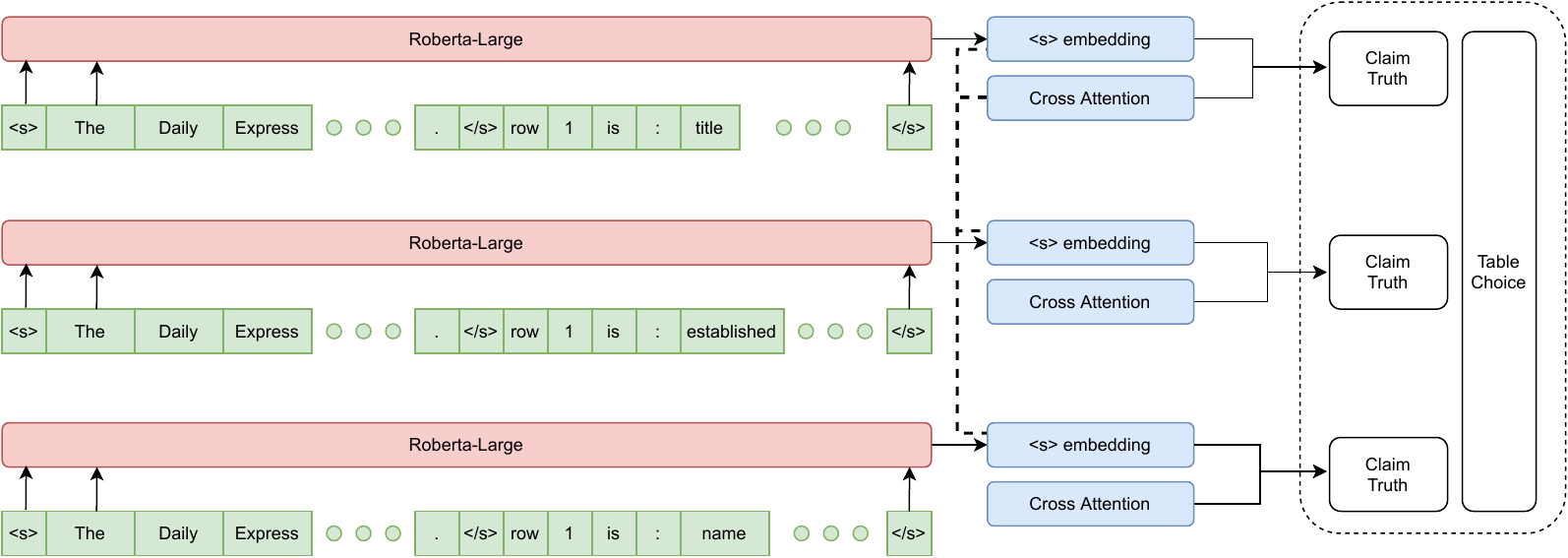}
    \caption{A diagram of our model, using the joint reranking- and verification approach described in Section~\ref{section:loss}. Linearised tables are encoded separately with RoBERTa. Then, cross-attention is used to contextualize each individual table with respect to the others. Finally, the model jointly predicts truth value and table selection.}
    \label{fig:model_overview}
\end{figure*}

Here, we investigate fact verification over tables in the open setting. We take inspiration from similar work on unstructured data~\citep{chen2017reading, nie2019combining, karpukhin2020dense, lewis2020retrieval}, proposing a two-step model which combines ad-hoc retrieval with a neural reader. 
Drawing on preliminary work in open question answering over tables~\citep{sun2016table}, we perform retrieval based on simple heuristic modeling of individual table cells. We combine this retriever with a RoBERTa-based~\citep{liu2019roberta} joint reranking-and-verification model, performing fusion of evidence documents in the verification component. This corresponds to the approach suggested for question answering by e.g.~\citet{izacard2020leveraging}.

We evaluate our models using the recently introduced TabFact dataset~\citep{chen2019tabfact}. While initially developed for the closed domain, the majority of claims are sufficiently context-independent that they can be understood without knowing which table they were constructed with reference to. As such, the dataset is suitable for the open domain as well. 
Our models represent a first step into the open domain, achieving open-domain performance exceeding the previous closed-domain state of the art---outside of \citet{eisenschlos-etal-2020-understanding}, which includes pretraining on additional synthetic data. 
We demonstrate significant gains from including multiple tables, and these gains are 
increasing as more tables are used. We furthermore present results using a more realistic setting where tables are retrieved not just from the 16,573 TabFact tables, but from the full Wikipedia dump. Our contributions can be summarized as follows:

\begin{enumerate}
    \item We introduce the first model for open-domain table fact verification, demonstrating strong performance exceeding the previous closed-setting state of the art.
    \item We propose two strategies with corresponding loss functions for modeling table fact verification in the open setting, suitable respectively for high verification accuracy or identifying if appropriate information has been retrieved for verification.
    \pagebreak
    \item In addition to our open-domain performance, our model achieves a new closed-domain state-of-the-art result.
    \item We report the first results on Wikipedia-scale open-domain table fact verification, using all tables from a Wikipedia dump as the backend.
\end{enumerate}
We release the source code for our experiments at \url{https://github.com/facebookresearch/OpenTableFactChecking}.

\section{Open-Domain Table Fact Verification}

Formally, the open table fact verification problem can be described as follows. Given a claim $q$ and a collection of tables $T$, the task is to determine whether $q$ is true or false. As such, we approach the task by modeling a binary verdict variable $v$ as $p(v|q,T)$. This is in contrast to the closed setting, where a single table $t_q \in T$ is given, and the task is to model $p(v|q,t_q)$. Since there are large available datasets for the closed setting~\citep{chen2019tabfact, gupta-etal-2020-infotabs}, it is reasonable to expect to exploit $t_q$ during training; however, at test time, this information may not be available. We follow a two-step methodology that is often adopted in open-domain setting for unstructed data~\citep{chen2017reading, nie2019combining, karpukhin2020dense, lewis2020retrieval} to our setting. Namely, given a claim query $q$, we retrieve a set of evidence tables $D_q \subset T$ (Section~\ref{section:retrieval}), and subsequently model $p(v|q,D_q)$ in place of $p(v|q,T)$ (Section~\ref{section:model}). 

\section{Entity-based Retrieval}
\label{section:retrieval}

We first design a strategy for retrieving an appropriate subset of evidence tables for a given query. For question answering over tables, \citet{sun2016table} demonstrated strong performance on retrieving relevant tables using entity linking information, following the intuition that many table cells contain entities. We take inspiration from these results. In their setting, claim entities are linked to Freebase entities, and string matching on the alias list is used to map entities to cells. To avoid reliance on a knowledge graph, we instead use only the textual string from the claim to represent entities, and perform approximate matching through dot products of bi- and tri-gram TF-IDF vectors.

We pre-compute bi- and trigram TF-IDF vectors $z(c^1_t), ..., z(c^m_t)$ for every table $t \in T$ with cells $c_t^1, ..., c_t^m$. Then, we identify the named entities $e^1_q, ..., e^n_q$ within the query $q$. For our experiments, we use the named entity spans for TabFact, provided by \citet{chen2019tabfact} as part of their LPA-model.\footnote{In the absence of named entity tags, named entity spans would first need to be found though an off-the-shelf named entity recognizer, such as SpaCy~\citep{spacy}.} 
We compute bi- and trigram TF-IDF vectors $z(e^1_q), ..., z(e^n_q)$ for the surface forms of those entities.  
To retrieve $D_q$ given $q$, we then score every $t \in T$. Since we are approximating entity linking between claim entities and cells, we let the score between an entity and a table be the \textit{best} match between that entity and any cell in the table. That is:
\begin{equation}\label{equation:entity_matching}
    score(q,t) = \sum\limits_{i = 1}^n \max\limits_{j = 1}^m z(e^i_q)^\intercal \cdot z(c^j_t)
\end{equation}
In other words, we compute for every entity the best match in the table, and score the table as the sum over the best matches. To construct the set of evidence tables $D_q$, we then retrieve the top-$k$ highest scoring tables. Our choice to use bi- and tri-gram TF-IDF as the retrieval strategy was determined empirically --- see Section~\ref{section:retrieval_performance} and Table~\ref{table:retrieval_performance} for experimental comparisons.

\section{Neural Verification}
\label{section:model}
To model $p(v|q, D_q)$, we employ a RoBERTa-based~\citep{liu2019roberta} late fusion strategy (see Figure~\ref{fig:model_overview} for a diagram of our model). Given a query $q$ with a ranked list of $k$ retrieved tables $D_q = (d_q^1, ..., d_q^k)$, we begin by linearising each table. Our linearisation scheme follows \citet{chen2019tabfact}. We first perform sub-table selection by excluding columns not linked to entities in the query. Here, we reuse the entity linking obtained during the retrieval step (see Section \ref{section:retrieval}), and retain only the three columns in which cells received the highest retrieval scores. We linearise each row separately, encoding entries and table headers. 
Suppose $r$ is a row with cell entries $c_1, c_2, ..., c_m$ in a table, where the corresponding column headers are $h_1, h_2, ..., h_m$. Row number $r$ is mapped to \textit{``row $r$ is : $h_1$ is $c_1$ ; $h_2$ is $c_2$; ... ; $h_m$ is $c_m$ ."} 

We construct a final linearisation $L_{q,t}$ for each query-table pair $q,t$ by prepending the query to the filtered table linearisation. We then encode each $L_{q,t}$ with RoBERTa, and obtain a contextualised embedding $f(d_q^k) \in \mathbb{R}^{n}$ for every table by using the final-layer embedding of the CLS-token. We construct the sequence of embeddings $f(d_q^1), ... f(d_q^k)$ for all $k$ tables.

When the model attempts to judge whether to rely on a given table for verification, other highly-scored tables represent useful contextual information (e.g., in the example in Figure~\ref{fig:example_query}, newspapers belonging to the same owner may be likely to also share political leanings). Nevertheless, each table embedding $f(d_q^k)$ is functionally independent from the embeddings of the other tables. As such, contextual clues from other tables cannot be taken into account. To remedy this, we introduce a cross-attention layer between all tables corresponding to the same query. We collect the embeddings $f(d_q^k)$ of each table into a tensor $F(D_q)$. We then apply a single multi-head self-attention transformation as defined by \citet{vaswani2017attention} to this tensor, and concatenate the result. That is, we compute an attention score for head $h$ from table $i$ to table $j$ with query $q$ as:
\begin{equation}
    \alpha^h_{ij} = \sigma\left(\frac{W^h_Q f(d_q^i) (W^h_K f(d_q^j))^T}{\sqrt{dim(K)}}\right)
\end{equation}
where $\sigma$ is the softmax function, and $W_Q$ and $W_K$ represent linear transformations to queries and keys, respectively. We then compute an attention vector for that head as:
\begin{equation}
    A_i^h = \sum\limits_{j \in D_q} \alpha_{ij} W^h_V f(d_q^j)
\end{equation}
and finally construct contextualized table representations through concatenation as:
\begin{equation}
    f^*(d_q^k) = [f(d_q^k), A_i^1, ..., A_i^h]
\end{equation}
We subsequently use $F^*(D_q)$, i.e. the tensor containing $f^*(d_q^1), ..., f^*(d_q^k)$, for downstream predictions. We note that our approach can be viewed as an extension of the Table-BERT algorithm introduced by~\citet{chen2019tabfact} to the multi-table setting, using an attention function to fuse together the information from different tables.

\begin{table*}[htb]
\centering
\begin{tabular}{l|rrrrrrr}
\toprule
Retrieval method      & H@1  & H@3  & H@5  & H@10 \\ \midrule
Query-matching word-level TF-IDF          & 41.7 & 54.2 & 59.0 & 65.3 \\
Query-matching character-level (2,3)-gram TF-IDF          & 34.7 & 45.5 & 50.2 & 56.8 \\\midrule
Entity-matching word-level exact match          & 48.2 & 57.9 & 64.2 & 67.3 \\
Entity-matching word-level TF-IDF          & 56.0 & 65.6 & 74.1 & 81.2 \\
Entity-matching character-level (2,3)-gram TF-IDF & \textbf{69.6} & \textbf{78.8} & \textbf{82.3} & \textbf{86.6} \\
Entity-matching character-level (1,2,3)-gram TF-IDF & 62.3 & 75.2 & 80.1 & 86.1 \\ \bottomrule
\end{tabular}
\caption{Retrieval accuracy for our entity-based TF-IDF retrieval along with several baselines for the TabFact validation set, computed using all 16,573 TabFact tables. We experiment with matching the entire query against table cells (above), and matching individual entities in the query against table cells using Equation~\ref{equation:entity_matching} (below). For all subsequent experiments we rely on character-level (2,3)-grams with entity-matching for retrieval.}
\label{table:retrieval_performance}
\end{table*}

\subsection{Training \& Testing}
\label{section:loss}

Relying on a closed-domain dataset provides a table with appropriate information for answering each query; namely, the table against which the claim is to be checked in the closed setting. Although this information is not available at test time, we can construct a training regime that allows us to exploit it to improve model performance. 
We experiment with two different strategies: jointly modeling reranking of tables along with verification of the claim, and modeling for each table a ternary choice between indicating truth, falsehood, or giving no relevant information. 
Later, we demonstrate how the former leads to increased performance on verification, while the latter gives access to a strong predictor for cases where no appropriate table has been retrieved.
\ignore{
We experiment with two different strategies: (1) \emph{jointly} modeling reranking of tables along with verification of the claim, and (2) modeling for each table a \emph{ternary} choice between indicating truth, falsehood, or giving no relevant information. 
Later, we demonstrate how the former leads to increased performance on verification, while the latter gives access to a strong predictor for cases where no appropriate table has been retrieved.
}
\ignore{
We experiment with two different strategies: \emph{joint reranking and verification} and \emph{ternary verification}. The former jointly models reranking of tables along with verification of the claim, and the latter models for each table a ternary choice between indicating truth, falsehood, or giving no relevant information. Later, we demonstrate how joint reranking and verification leads to increased performance on verification, while ternary verification gives access to a strong predictor for cases where no appropriate table has been retrieved.
}

\paragraph{Joint reranking and verification}

For the joint reranking  and verification approach, we assume that a \textit{best} table for answering each query is given and can be used to learn a ranking function. We model this as selecting the right table from $D_q$, e.g., through a categorical variable $s$ that indicates which table should be selected. We then learn a joint probability of $s$ and the truth value of the claim $v$ over the tables for a given query. Assuming that $s$ and $v$ are independent, $p(s, v | q, D_q)$ is also a categorical distribution with one correct outcome that can be optimized for (that is, one correct pair of table and truth value). As such, we let:
\begin{equation}
    p(s, v | q, D_q) = \sigma(W( F^*(D_q)_s)_v)
\end{equation}
Where $W: \mathbb{R}^{2n} \to \mathbb{R}^2$ is an MLP and $\sigma$ is the softmax function. At train time, we obtain one cross-entropy term corresponding to $p(s, v | q, D_q)$ per query. At test time, we marginalize over $s$ to obtain a final truth value:
\begin{equation}\label{equation:ternary_table}
    p_v(v | q, D_q) = \sum\limits_{t \in D_q} p(v, s=t | q, D_q)
\end{equation}\label{equation:ternary_expectation}
This formulation has the additional benefit of also allowing us to make a prediction on which table matches the query. We can do so by marginalizing over $v$:
\begin{equation}\label{equation:reranking}
    p_s(s | q, D_q) = \sum\limits_{v_q \in \{true,false\}} p(s, v=v_q | q, D_q)
\end{equation}
With this loss, we train the model by substituting for $D_q$ a set $D^*_q$ containing wherein the gold table is guaranteed to appear. We ensure this by replacing the lowest-scored retrieved table in $D_q$ with the gold table whenever it has not been retrieved.

\paragraph{Ternary verification}

At test time, there may be cases where a table refuting or verifying the fact is not contained in $D_q$. For some applications, it could be useful to identify these cases. We therefore design an alternative variant of our system better suited for this scenario. Intuitively, each table can represent three outcomes -- the query is true, the query is false, or the table is irrelevant. We can model this through a ternary variable $i$ such that for table $t$:
\begin{equation}
    p(i | q, t, D_q) = \sigma(W^\prime(F^*(D_q)_t)_{i})
\end{equation}
Where $W^\prime: \mathbb{R}^{2n} \to \mathbb{R}^3$ is an MLP and $\sigma$ is the softmax function. During training, we assign \textit{true} or \textit{false} to the gold table depending on the truth of the query, and \textit{irrelevant} to every other table. We then use the mean cross-entropy over the tables associated with each query as the loss for each example. At test time, we compute the truth value $v$ of each query as:
\begin{equation}
    \sum\limits_{t \in D_q} p(i = true | q, t) > \sum\limits_{t \in D_q} p(i = false | q, t)
\end{equation}

\begin{table*}[htb]
\centering
\begin{tabular}{l|rrrrr}
\toprule
Model                     & Dev  & Test & Simple Test & Complex Test & Small Test \\ \midrule
Table-BERT \citep{chen2019tabfact} & 66.1 & 65.1 & 79.1        & 58.2         & 68.1       \\
LogicalFactChecker \citep{zhong2020logicalfactchecker} & 71.8 & 71.7 & 85.4 & 65.1 & 74.3 \\
ProgVGAT \citep{yang-etal-2020-program} &  74.9 & 74.4 & 88.3 & 67.6 & 76.2 \\
TAPAS \citep{eisenschlos-etal-2020-understanding}* &  81.0 & 81.0 & 92.3 & 75.6 & 83.9 \\ 
Ours (Oracle retrieval)         & 78.2 & 77.6 & 88.9        & 72.1         & 79.4       \\ \midrule
Ours (1 retrieved table)         & 74.1 & 73.2 & 86.7 & 67.8 &	76.6       \\  
Ours (Ternary loss, 3 tables)           & 73.8 & 73.5 & 86.9 & 68.1 & 76.9        \\
Ours (Ternary loss, 5 tables)           & 74.1 & 73.7 & 87.1 & 67.9 & 76.5       \\
Ours (Ternary loss, 10 tables)           & 73.9	& 73.1 & 86.5 & 67.9 & 77.3       \\
Ours (Joint loss, 3 tables)            & 74.6 & 73.8	& 87.0 & 68.3 &	78.1       \\
Ours (Joint loss, 5 tables)           & \textbf{75.9} & \textbf{75.1} & \textbf{87.8} & \textbf{69.5} & \textbf{77.8}       \\
Ours (Joint loss, 10 tables)          & 73.9 & 73.8	& 86.9 & 68.1 & 76.9 \\ 
\bottomrule
\end{tabular}
\caption{Prediction accuracy of our RoBERTa-based model on the official splits from the TabFact dataset. We include closed-domain performance of several models from the literature, as well as the performance of our model in both the closed and the open domain, using both proposed loss functions. The first section of the table contains closed-domain results, the second open-domain. * employs intermediary pretraining on additional synthetic data.
}
\label{table:tabfact_results}
\end{table*}

\section{Experiments}

We apply our model to the TabFact dataset~\citep{chen2019tabfact}, which consists of 92,283 training, 12,792 validation and 12,792 test queries over 16,573 tables. The task is binary classification of claims as true or false, with an even proportion of the two classes in each split. To benchmark our open-domain models and construct performance bounds, we begin by evaluating in the closed domain. As an upper bound, we can then compare against the performance of the closed-domain system scored using a single table retrieved through an oracle. As a lower bound, we can again use the closed-domain system, but using the highest-ranked table according to our TF-IDF retriever. The evaluation metric is simply prediction accuracy.

\subsection{Retrieval}
\label{section:retrieval_performance}

We choose bi- and tri-gram TF-IDF as the retrieval strategy empirically. To address the comparative performance of this choice, we compute and rank in Table~\ref{table:retrieval_performance} the retrieval scores obtained through our strategy on the TabFact test set. We compare against several alternative strategies: bi- and trigram TF-IDF vectors for all words in the query (rather than just the entities), word-level TF-IDF vectors for entities, and entity-level exact matching. Our bi- and tri-gram TF-IDF strategy yields by far the strongest performance. We furthermore demonstrate how the exclusion of unigrams from the TF-IDF vectors slightly increases performance.

\subsection{Verification}
\label{section:verification_results}

In Table~\ref{table:tabfact_results}, we compare our best-performing models to the closed-setting system from \citet{chen2019tabfact}, as well as to several recent models from the literature~\citep{zhong2020logicalfactchecker, yang-etal-2020-program, eisenschlos-etal-2020-understanding}. We include results with both losses as discussed in Section~\ref{section:model}, using varying numbers of tables. 

With an accuracy of $75.1\%$, we obtain the best open-domain results with our model using the joint reranking-and-verification loss and five tables. We see performance improvements when increasing the number of tables, both from 1 to 3 and from 3 to 5. 
In the closed domain, the $77.6\%$ accuracy our model achieves is a significant improvement over the $74.4\%$ the strongest comparable baseline reached. This may be due to our use of RoBERTa, which has previously been found to perform well for linearised tables~\citep{gupta-etal-2020-infotabs}.

Relying purely on TF-IDF for retrieval --- that is, using our system with only one retrieved table --- yields a performance of $73.2\%$. This is a surprisingly small decrease compared to the closed domain, given that an incorrect table is provided in approximately a third of all cases (see Table~\ref{table:retrieval_performance}). We suspect that many cases for which the retriever fails are also cases for which the closed-domain model fails. To make sure we are not seeing the effect of false negatives (e.g., tables which are not the gold table, but which nevertheless have the information to verify the claim), we run the model in a setting where one retrieved table is used, but the gold table is removed from the retrieval results; here, the model achieves an accuracy of only $56.2\%$. We furthermore test a system relying on a random table rather than a retrieved table; with a performance drop to $53.1$, we find that the information in the retrieved table is indeed crucial to obtain high performance (rather than the performance being purely a consequence of, say, RoBERTa weights).


\begin{table}[t]
\centering
\begin{tabular}{l|rrr}
\toprule
Model                          & R@1 & R@2-3 & R@4-5 \\ \midrule
Oracle retrieval              & 80.6 & 74.1 & 75.0 \\ \midrule
1 table               & 80.6 & 55.6 & 53.9  \\
3 tables                & 78.8 & 66.7 & 58.2 \\
5 tables                & 79.4 & 73.1 & 71.7 \\\bottomrule
\end{tabular}
\caption{Peformance of our RoBERTa-based model on the parts of the TabFact test set where our TF-IDF retriever assigns the gold table rank respectively 1, 2-3, or 4-5.}
\label{table:gold_rank_performance}
\end{table}

To understand how our model derives improvement from the addition of more tables, we compute in Table~\ref{table:gold_rank_performance} the performance of our reranking-and-verification model when TF-IDF returns the correct table at rank 1, rank 2-3, or rank 4-5. Immediately, we notice a much stronger improvement from using multiple tables when TF-IDF fails to correctly identify the gold table. This is natural, as those are exactly the cases where our model (as opposed to the baseline) has access to the appropriate information to verify or refute the claim.

Interestingly, using three tables improves on using one table even when the gold table is not included among the top three (from $53.9\%$ to $58.2\%$), and using five tables improves on using three tables also when the gold table \textit{is} included among the top three (from $66.7\%$ to $73.1\%$). Manual inspection reveals that our model in some cases relies on correlations between tables
--- if a sports team loses games in three tables, then that may give a higher probability of that team also losing in an unretrieved, hypothetical fourth table. To test this, we apply the model in a setting where we retrieve the top five tables \textit{excluding} the gold table, and a setting where we use five random tables. Using highly scored (but wrong) tables, we achieve a performance of $59.4\%$, a significant improvement on the $53.1\%$ we achieve using random tables. This supports our hypothesis that other \textit{good} tables can provide useful background context for verification. 

It should be noted that such inferences, while increasing model performance, may also increase the degree to which the model exhibits biases. Depending on the application, this may as such not be a desirable basis for verification. Returning to the example in Figure~\ref{fig:example_query}, inferring ownership on the basis of political affiliation when no other information is available may increase accuracy \textit{on average}, but it can also lead to erroneous or biased decisions (indeed, for the claim in the example, the prediction would be wrong).

\subsection{Ablation Tests}

Our best-performing model from Table~\ref{table:tabfact_results} relies on two innovations: The cross-attention function which contextualizes retrieved tables in relation to each other, and the joint reranking-and-verification loss. In Table~\ref{table:ablation}, we evaluate the model without either of these. Leaving the attention function out is simple --- we use $f(d_q^k)$ for each table directly for predictions. We model performance without the reranking component of our loss function by assuming a uniform distribution over the tables.

\begin{table}[t]
\centering
\begin{tabular}{l|r}
\toprule
Model             & Accuracy \\ \midrule
Full model        & 75.1                         \\
- Attention       & 73.6                         \\
- Joint objective & 72.9                         \\
- Both            & 71.2                         \\ \bottomrule
\end{tabular}
\caption{Ablation study for our model, performing verification with the five-table version on the TabFact test set. We remove respectively our cross-attention function, the reranking component in the loss, and both.}
\label{table:ablation}
\end{table}

As can be seen, the combination of both is strictly necessary to obtain strong performance --- indeed, without our joint objective, the model performs worse than simply applying the baseline model to the top table returned by TF-IDF as in Table~\ref{table:tabfact_results}. 
The ability for the model to express the relative relatedness of tables to the query is crucial. We include further investigation of the role our cross-attention mechanism plays in Appendix~\ref{appendix:attention}.

\subsection{Predicting Insufficient Information}
\label{section:insufficient}

In realistic settings, some claims will not be directly answerable from any retrieved table. 
In such cases, it can be valuable to explicitly inform the user --- giving false verifications or refutations when sufficient information is not available is misleading, and can decrease user trust. 
To model a scenario where the lack of relevant information must be detected, we create a classification task wherein the model must predict for all examples, whether the gold table is among the $k$ documents in $D_q$. 

Using the ternary loss, our model directly gives the probability of each table containing appropriate information as $(1 - p(I_t=irrelevant|q,t))$. We can estimate the suitability of the best retrieved table for verifying the claim as $\max\limits_t (1 - p(I_t=irrelevant|q,t))$, and apply a threshold $\tau_1$ to classify $D_q$ as suitable or unsuitable. For the joint loss, a more indirect approach is necessary. Intuitively, if our model is too uncertain about which table answers the query, there is a high likelihood that no suitable table has been retrieved. This corresponds to the entropy of the reranking component $H_s(s|q,D_q)$ after marginalizing over the truth value of the claim exceeding some threshold $\tau_2$.

\begin{figure}[t]
\centering
\begin{tikzpicture} 
\begin{axis}[xlabel=Recall,
ylabel=Precision,
scale=0.75,
xmin=0,
xmax=1,
ymin=0.8,
ymax=1,
legend style={font=\fontsize{9}{11}\selectfont}] 
\addplot[color=red,mark=triangle*] table[x=jointr,y=jointp] {data/pr.dat};
\label{table:threeway_classification:p1}
\addplot[color=blue,mark=x] table[x=terr, y=terp] {data/pr.dat}; 
\label{table:threeway_classification:p2}
\addplot[mark=none, black,thick, dotted] coordinates {(0,0.82) (1,0.82)};
\label{table:threeway_classification:p3}
\end{axis}%
\end{tikzpicture}
\caption{Precision-recall curve for determining whether a set of five retrieved tables in the TabFact validation set contains the gold table, using respectively entropy of the reranking scores with our joint loss (\ref{table:threeway_classification:p1}) or the maximum probability of some table being the gold table with our ternary loss, (\ref{table:threeway_classification:p2}). We also include a most frequent class baseline (\ref{table:threeway_classification:p3}).}\label{table:threeway_classification}
\end{figure}
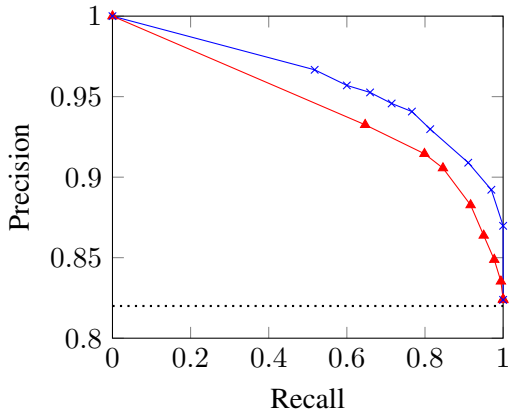

We compare these strategies in Figure~\ref{table:threeway_classification}, obtaining Precision-Recall curves by measuring at varying $\tau_1$ and $\tau_2$. We find that while both approaches outperform a most frequent class baseline by a significant margin, the ternary loss performs better than the joint loss. As such, the choice between the two losses represents a tradeoff between raw performance (see Tables~\ref{table:tabfact_results} and~\ref{table:tabfact_results_fullwiki}) and the ability to identify missing or incomplete information.

\subsection{Wikipedia-scale Table Verification}

In our experiments so far, we have relied on the 16,573 TabFact tables as the knowledge source. The tables selected for TabFact were taken from WikiTables~\citep{bhagavatula2013methods}, and filtered so as to exclude ``overly complicated and huge tables"~\citep{chen2019tabfact}. Moving beyond the scope of that dataset, a fully open fact verification system should be able to verify claims over even larger collections of tables --- for example, the full set of tables available on Wikipedia. To make a preliminary exploration of that larger-scale setting, we include in Table~\ref{table:tabfact_results_fullwiki} the performance of our approach evaluated using roughly 3 million tables automatically extracted from Wikipedia.

\begin{table}[h!]
\centering
\begin{tabular}{l|r}
\toprule
Model             & Accuracy \\ \midrule
RoBERTa only        & 52.1                         \\
Ours (1 table)       & 53.6                         \\
Ternary loss, 3 tables & 55.8                         \\
Ternary loss, 5 tables            & 57.5                         \\
Joint loss, 3 tables & 56.1 \\
Joint loss, 5 tables            & \textbf{58.1}                         \\ \bottomrule
\end{tabular}
\caption{Performance of our RoBERTa-based model on the TabFact test set, using all Wikipedia tables rather than just the TabFact tables as a backend.}
\label{table:tabfact_results_fullwiki}
\end{table}

As can be seen, our approach improves on the naive strategy of using a single table and a closed-domain verification component also in this more complex setting. To verify that the inference happens on the basis of the retrieved tables and not simply the RoBERTa-weights, we include also the performance of a model which simply uses classification on top of a RoBERTa-encoding of the claim. Similar to our previous experiments, the joint-loss model with five retrieved tables performs the strongest. We note that it is unclear whether the performance we observe here originates from correlations obtained through background information (as we see in Section~\ref{section:verification_results} when the retriever fails to find the appropriate table), or due to verification against a single entirely appropriate table happening at a lower rate than when using TabFact.

\section{Related Work}

Semantic querying against large collections of tables has previously been studied for question answering. \citet{sun2016table} used string matching between aliases of linked entities to search millions of tables crawled from the Web, with retrieved table cells providing evidence for a question answering task. \citet{jauhar2016tables} demonstrated strong results with a Lucene index and a Markov Logic Network-based model for answering scientific questions. Recently, \citet{chakrabarti2020open, chakrabarti2020tableqna} developed an improved model for table retrieval combining neural representations of the table and the query with a BM25 index.

\citet{cafarella2008webtables, cafarella2009data} employed keyphrase-based table retrieval by reranking a list of tables returned by a search engine. \citet{pimplikar2012answering} used a graphical model to perform retrieval on the basis of co-occurence statistics, table metadata, and column headers. In \citep{ghasemi2018tabvec}, non-parametric clustering was employed as a strong heuristic for table retrieval. \citet{zhang2018ad} introduced a ranking method based on mapping available features into several semantic spaces. Recently, \citet{zhang2019table2vec} introduced a neural method for table retrieval and completion using word- and entity-embeddings of table elements.

Neural modeling of tables has been the subject of several recent papers. Aside from the original BERT-based model in~\citep{chen2019tabfact}, the closest to our work is~\citep{yin2020tabert}. In these paper, a pretrained BERT-based encoder for tables is introduced and demonstrated to yield strong improvements on several semantic parsing tasks. \citet{chen2019colnet} introduced a model to automatically predict and compare column headers for tables in order to find semantically synonymous schema attributes. Similarly, \citet{zhang2019auto} introduced an autoencoder for predicting table relatedness. 

Closed-domain semantic parsing over tables has been studied extensively in the context of question answering (e.g., \citet{pasupat-liang-2015-compositional, khashabi2016question, yu-etal-2018-spider}). In~\citet{zhong2020logicalfactchecker}, a logic-based fact verification system was introduced to improve on the model presented in the initial TabFact paper \citep{chen2019tabfact}. \citet{yang-etal-2020-program} builds on the program induction model also introduced in \citet{chen2019tabfact}, using a graph neural network to verify generated programs. Orthogonally, a similar dataset for table-based natural language inference was introduced by \citet{gupta-etal-2020-infotabs} --- interestingly, like in our experiments, they found RoBERTa-large to work extremely well for linearised tables. Finally, \citet{herzig-etal-2020-tapas, eisenschlos-etal-2020-understanding} introduced BERT-based models for various table semantic tasks, extending BERT with additional position embeddings denoting columns and rows.

Open-domain fact verification and question answering over unstructured, textual data has been studied in a series of recent papers. Early work resulted in several highly sophisticated full pipeline systems \citep{brill2002analysis, ferrucci2010building, sun2015open}. These provided inspiration for the influential DrQA model \citep{chen2017reading}, which like ours relies on a TF-IDF-based heuristic retrieval model, and a complex reading model. Recent work~\citep{karpukhin2020dense, lewis2020retrieval} has built on this approach, developing learned dense retrieval models with dot-product indexing~\citep{johnson2017faiss}, and increasingly advanced pretrained transformer-models for reading. The development of similarly fast, reliable and learnable indexing techniques for tables as well as text is an important direction for future work.

Concurrently with our work, \citet{chen2020open} have introduced a BERT-based model to perform question answering over open collections of data including tables. Like ours, their model consists of separate retriever- and reader-steps. Their best-performing reader employs a long-range sparse attention transformer~\citep{ravula2020etc} to jointly summarize all retrieved data. As in our case, their model demonstrates significant improvements from using multiple retrieved tables.

\section{Conclusion}

We have introduced a novel model for fact verification over large collections of tables, along with two strategies for exploiting closed-domain datasets to increase performance. Our approach performs on par with the current closed-domain state of the art, with larger gains the more tables we include. When using an oracle to retrieve a reference table, our approach also represents a new closed-domain state of the art. Finally, we have made an initial foray into Wikipedia-scale open-domain table fact verification, demonstrating improvements from multiple tables also when using a full set of Wikipedia tables as the knowledge source. Our results indicate that the use of multiple tables can provide contextual clues to the model even when those tables do not explicitly verify or refute the claim, because they can provide evidence for the \textit{probability} of the claim. This is a double-edged sword, as reliance on such clues can increase performance while also inducing biased claims of truthfulness. Care will be needed in future work to disentangle the positive and negative aspects of this phenomenon.

\section*{Acknowledgments}

We would like to thank Fabio Petroni and Nicola De Cao for helpful discussions and comments.

\bibliography{anthology, others_rebibered}
\bibliographystyle{acl_natbib}

\appendix
\section{Performance using RoBERTa-base}
\label{appendix:roberta_base}

In our paper, we have reported results using the larger version of RoBERTa with additional hidden layers and greater dimensionality. \citet{liu2019roberta} also include a smaller version, RoBERTa-base, corresponding to the BERT-base model of~\citet{devlin-etal-2019-bert}. In Table~\ref{table:tabfact_results_base}, we report results corresponding to those of Table~\ref{table:tabfact_results} for our joint model using RoBERTa-base instead of RoBERTa-large.

Interestingly, the performance gain from using multiple tables is even larger for RoBERTa-base (an increase of $4.1$ rather than $1.9$ point accuracy from using five tables, for example). One explanation could be that some information necessary to verify certain facts may be encoded in the weights of the larger RoBERTa. We attempt to investigate this using random tables; however, with five random tables, RoBERTa-large and RoBERTa-base reach an almost equivalent respective performance of $52.1$ and $52.0$. As such, we believe that RoBERTa-large exploits the correlations between tables which we discuss in Section~\ref{section:verification_results} better than RoBERTa-base.

\section{Hyperparameters}
\label{appendix:hyperparameters}

Our model uses RoBERTa~\citep{liu2019roberta} to encode each table into vectors. On top of RoBERTa we employ key-value self-attention~\citep{vaswani2017attention} with two attention heads. We then use an MLP consisting of a linear transformation to $h=3072$ hidden units, followed by $tanh$-activation and linear projection to the output space. During training, we employ dropouts with probability $0.1$ before each linear transformation in the MLP. The hyperparameters for all experiments were selected using the TabFact development set (with TabFact tables as the backend).

\begin{table*}[htb]
\centering
\begin{tabular}{l|rrrrr}
\toprule
Model                     & Dev  & Test & Simple\\\midrule 
Table-BERT \citep{chen2019tabfact} & 66.1 & 65.1 & 79.1        & 58.2         & 68.1       \\
LogicalFactChecker \citep{zhong2020logicalfactchecker} & 71.8 & 71.7 & 85.4 & 65.1 & 74.3 \\
ProgVGAT \citep{yang-etal-2020-program} &  74.9 & 74.4 & 88.3 & 67.6 & 76.2 \\
TAPAS \citep{eisenschlos-etal-2020-understanding}* &  81.0 & 81.0 & 92.3 & 75.6 & 83.9 \\ 
Ours (Roberta-base, oracle retrieval)         & 72.3	& 69.9 & 85.7 & 62.1 & 71.4       \\
Ours (Roberta-large, oracle retrieval)         & 78.2 & 77.6 & 88.9        & 72.1         & 79.4       \\ \midrule 
Ours (Roberta-base, 1 table)            & 64.7 & 64.2 & 78.4 & 57.8 & 67.3       \\
Ours (Roberta-base, 3 tables)           & 67.8 & 67.0 &	81.5 & 60.5 & 69.4        \\
Ours (Roberta-base, 5 tables)           & 68.1 & 68.3 & 83.4 & 61.2 & 70.2       \\
Ours (Roberta-base, 10 tables)           & 67.5	& 67.1 & 82.3 & 59.9 & 70.1       \\
\midrule
Ours (Roberta-large, 1 tables)           & 74.1 & 73.2 & 86.7 & 67.8 &	76.6        \\
Ours (Roberta-large, 3 table)            & 74.6 & 73.8	& 87.0 & 68.3 &	78.1       \\
Ours (Roberta-large, 5 tables)           & 75.9 & 75.1 & 87.8 & 69.5 & 77.8       \\
Ours (Roberta-large, 10 tables)          & 73.9 & 73.8	& 86.9 & 68.1 & 76.9 \\ 
\bottomrule
\end{tabular}
\caption{Prediction accuracy of our RoBERTa-based model on the official splits from the TabFact dataset, using RoBERTa-base in addition to RoBERTa-large. The first section of the table contains closed-domain results, the second and third open-domain. All results use the joint objective described in Section~\ref{section:loss}. * employs intermediary pretraining on additional synthetic data.}
\label{table:tabfact_results_base}
\end{table*}

We train the model using Adam~\citep{kingma2014adam} with a learning rate of $5e-6$. We use a linear learning rate schedule, warming up over the first $30000$ batches. We use a batch size of $32$. Training was done on 8 NVIDIA Tesla V100 Volta GPUs (with 32GB of memory), and completed in approximately 36 hours.

\section{Retrieval accuracy for TabFact splits}
\label{appendix:acc_splits}

The TabFact dataset comes with several different data splits. We include here the performance of our retrieval component for each split:

\begin{table}[h]
\centering
\begin{tabular}{l|rrrrrrr}
\toprule
Dataset      & H@1  & H@3  & H@5  & H@10 \\ \midrule
Train        & 59.5 & 71.2 & 74.8 & 79.2 \\
Dev          & 69.6 & 78.8 & 82.3 & 86.6 \\
Test & 69.7 & 78.7 & 81.9 & 86.3 \\ 
Simple Test  & 92.7 & 97.1 & 98.1 & 99.0 \\
Complex Test & 64.7 & 75.2 & 79.5 & 84.8 \\
Small Test   & 82.1 & 89.6 & 91.4 & 94.7 \\ \bottomrule
\end{tabular}
\caption{Retrieval accuracy with our entity-based TF-IDF heuristic on the different TabFact splits.}
\label{table:retrieval_accuracy}
\end{table}

\section{Reranking Performance}

In Section~\ref{section:model}, we introduced our model as a joint system for fact verification and evidence reranking. A benefit of our formulation is the ability to reason about the ability of our model to rerank by marginalizing over the truth value of the claim, following Equation~\ref{equation:reranking}. In Table~\ref{table:performance_reranking}, we compare the table retrieval ranking performance of our joint model to a model only trained for reranking, as well as to the TF-IDF baseline.

\begin{table}[h]
\centering
\begin{tabular}{l|rrr}
\toprule
Model                         & H@1  & H@3  & H@5 \\ \midrule
TF-IDF                        & 69.6 & 78.8 & 82.3    \\
Reranking only                & 69.9 & 78.9  & 82.3     \\
Ours (no attention) & 67.4 & 78.3 & 82.3    \\
Ours (attention)    & 70.9 & 79.4     & 82.3    \\ \bottomrule
\end{tabular}
\caption{Ranking performance on the TabFact validation set, using either our TF-IDF retriever alone or reranking with our model. We test a version of our model using only a reranking loss, as well as joint-loss model with and without attention.}
\label{table:performance_reranking}
\end{table}

As can be seen, our joint loss provides a slight performance improvement when the attention component is included. Interestingly, the joint-loss model performs better than a system trained purely for reranking --- this highlights the complementary nature of the reranking and verification tasks.

\section{The Role of Attention}
\label{appendix:attention}

\def\classNames{{"1","2","3","4","5"}} 
\definecolor{scaleblue}{rgb}{0.2, 0.4, 1.0}

\def\numClasses{5} 

\begin{figure*}[t]
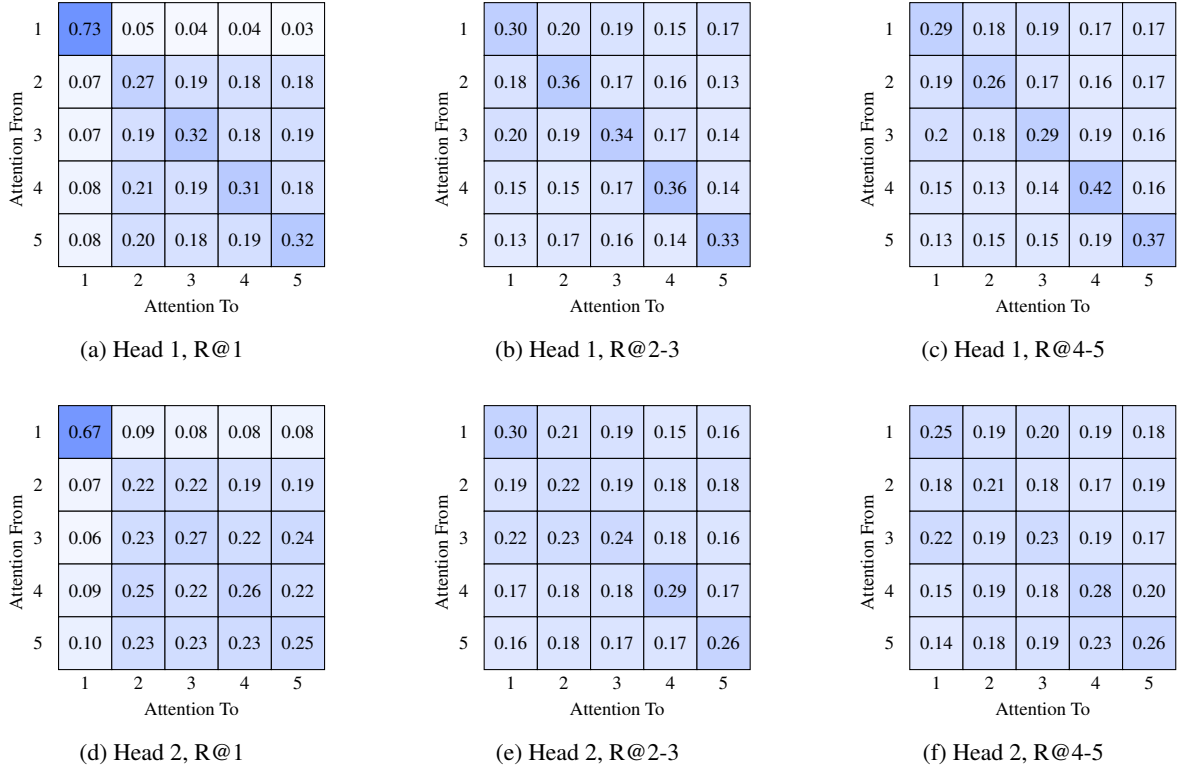

    \centering
    \def\myScale{0.7}
    
    \begin{subfigure}[b]{0.3\textwidth}
    \centering
    \begin{tikzpicture}[
    scale = \myScale,
    font={\scriptsize},
    ]

\def\myConfMat{{
{0.73,	0.05,	0.04,	0.04,	0.03},
{0.07,	0.27,	0.19,	0.18,	0.18},
{0.07,	0.19,	0.32,	0.18,	0.19},
{0.08,	0.21,	0.19,	0.31,	0.18},
{0.08,	0.20,	0.18,	0.19,	0.32},
}}

\input{cm_inner}
\end{tikzpicture}
\caption{Head 1, R@1}
    \end{subfigure}
    \hfill
    \begin{subfigure}[b]{0.3\textwidth}
    \centering
    \begin{tikzpicture}[
    scale = \myScale,
    font={\scriptsize}, 
    ]

\def\myConfMat{{
{0.30,	0.20,	0.19,	0.15,	0.17},
{0.18,	0.36,	0.17,	0.16,	0.13},
{0.20,	0.19,	0.34,	0.17,	0.14},
{0.15,	0.15,	0.17,	0.36,	0.14},
{0.13,	0.17,	0.16,	0.14,	0.33},
}}

\input{cm_inner}
\end{tikzpicture}
\caption{Head 1, R@2-3}
    \end{subfigure}
    \hfill
    \begin{subfigure}[b]{0.3\textwidth}
    \centering
    \begin{tikzpicture}[
    scale = \myScale,
    font={\scriptsize}, 
    ]

\def\myConfMat{{
{0.29,	0.18,	0.19,	0.17,	0.17},
{0.19,	0.26,	0.17,	0.16,	0.17},
{0.2,	0.18,	0.29,	0.19,	0.16},
{0.15,	0.13,	0.14,	0.42,	0.16},
{0.13,	0.15,	0.15,	0.19,	0.37},
}}

\input{cm_inner}
\end{tikzpicture}
\caption{Head 1, R@4-5}
    \end{subfigure}%
    \vspace{0.5cm}
    
    \begin{subfigure}[b]{0.3\textwidth}
    \centering
    \begin{tikzpicture}[
    scale = \myScale,
    font={\scriptsize},
    ]

\def\myConfMat{{
{0.67,	0.09,	0.08,	0.08,	0.08},
{0.07,	0.22,	0.22,	0.19,	0.19},
{0.06,	0.23,	0.27,	0.22,	0.24},
{0.09,	0.25,	0.22,	0.26,	0.22},
{0.10,	0.23,	0.23,	0.23,	0.25},
}}

\input{cm_inner}
\end{tikzpicture}
\caption{Head 2, R@1}
    \end{subfigure}
    \hfill
    \begin{subfigure}[b]{0.3\textwidth}
    \centering
    \begin{tikzpicture}[
    scale = \myScale,
    font={\scriptsize}, 
    ]

\def\myConfMat{{
{0.30,	0.21,	0.19,	0.15,	0.16},
{0.19,	0.22,	0.19,	0.18,	0.18},
{0.22,	0.23,	0.24,	0.18,	0.16},
{0.17,	0.18,	0.18,	0.29,	0.17},
{0.16,	0.18,	0.17,	0.17,	0.26},
}}

\input{cm_inner}
\end{tikzpicture}
\caption{Head 2, R@2-3}
    \end{subfigure}
    \hfill
    \begin{subfigure}[b]{0.3\textwidth}
    \centering
    \begin{tikzpicture}[
    scale = \myScale,
    font={\scriptsize}, 
    ]

\def\myConfMat{{
{0.25,	0.19,	0.20,	0.19,	0.18},
{0.18,	0.21,	0.18,	0.17,	0.19},
{0.22,	0.19,	0.23,	0.19,	0.17},
{0.15,	0.19,	0.18,	0.28,	0.20},
{0.14,	0.18,	0.19,	0.23,	0.26},
}}

\input{cm_inner}
\end{tikzpicture}
\caption{Head 2, R@4-5}
    \end{subfigure}%

    \caption{Confusion matrices for the cross-attention between each pair of tables for the five-table version of our model. Each head is represented separately, and individual figures are included for the parts of the dataset where our TF-IDF retriever assigns the gold table rank respectively 1, 2-3, or 4-5.}
    \label{figure:attention}
\end{figure*}		

An interesting question is the role attention plays in our model. As can be seen from Tables \ref{table:tabfact_results} and \ref{table:performance_reranking}, our cross-attention module is necessary to achieve high performance -- without it, the model struggles to identify which table should be used for verification. To investigate the function of attention, we plot in Figure \ref{figure:attention} the strength of the cross-attention between each table for our five-table model. We produce separate plot for the two attention heads, as well as for each of the splits used in Table~\ref{table:gold_rank_performance} representing the parts of the dataset where our TF-IDF retriever assigns the gold table rank respectively 1, 2-3, or 4-5.

For both attention heads, the attention function has clearly distinct behaviour when the gold table is retrieved as top 1; the degree to which that table attends to itself is much greater. We suspect that this is because of ``easy" cases, where the attention function is used to separate a clearly identifiable ``appropriate" table from the other tables. In harder cases, the model uses the attention focus to compare information across tables. To test this, we run the model in a setting where four random tables are used along with the gold table. In that setting, the division is even clearer. For the gold table, respectively $86$ and $82$ percent of the attention for the two heads is on average focused on itself; for the four random tables, the attention is evenly distributed over all tables except the gold table. 

To distinguish the two heads, we in general see the first head exhibit a pattern of behaviour where each table assigns the majority of attention to itself --- especially when that table is the gold table. The second head seemingly encodes a more even spread over the retrieved tables, perhaps representing general context more than an attempt to identify the gold table.

\end{document}